\documentclass[]{FudanCVL}

\usepackage{silence}
\usepackage[misc]{ifsym}
\usepackage{xspace}

\newcommand{\methodname}{SAM-MT\xspace}

\title{\vspace{-1pt}{\fontsize{16pt}{19.5pt}\selectfont \methodname: Real-Time Interactive Multi-Target Video Segmentation}}

\author[1]{Ruiqi Shen}
\author[2]{Chang Liu}
\author[1]{Henghui Ding}

\affiliation[1]{Fudan University}
\affiliation[2]{Shanghai University of Finance and Economics}

\usepackage[utf8]{inputenc}
\usepackage[T1]{fontenc}
\usepackage{color}
\usepackage{epsfig}
\usepackage{graphicx}
\usepackage{booktabs}
\usepackage{tabularx}
\usepackage{multirow}
\usepackage{amsmath,amsfonts,amssymb}
\usepackage{bm}
\usepackage{nicefrac}
\usepackage{microtype}
\usepackage{xspace}
\usepackage{url}
\usepackage[table]{xcolor}
\usepackage{caption}
\usepackage{subcaption}
\usepackage{makecell}
\usepackage{cleveref}
\usepackage{stfloats}
\usepackage{xcolor}
\usepackage{wrapfig}
\definecolor{citecolor}{HTML}{0071BC}
\usepackage{hyperref}
\hypersetup{colorlinks=true, citecolor=citecolor, linkcolor=citecolor}
\usepackage{bbm}

\crefname{figure}{Fig.}{Figs.}
\crefname{table}{Tab.}{Tabs.}

\definecolor{defaultColor}{RGB}{230, 230, 250}

\definecolor{rred}{RGB}{245, 152, 153}
\definecolor{oorange}{RGB}{253, 205, 154}
\definecolor{yyellow}{RGB}{255, 240, 180}
\definecolor{ggreen}{RGB}{200, 230, 200}

\abstract{
Modern Video Object Segmentation (VOS) involves tracking and segmenting user-specified targets. While recent approaches have achieved remarkable performance in single-target scenarios, extending them to multi-target settings typically involves replicating the single-target processing for each individual object, resulting in reduced frame rates (FPS) with unbounded latency as target count increases. Built upon Segment Anything 2 (SAM2), we propose \textbf{SAM-MT}, which addresses this by transforming the model into an interactive framework for real-time \textbf{M}ulti-\textbf{T}arget video segmentation. SAM-MT uses explicit queries to represent different individual targets, in parallel with a shared representation for global context. It employs decoupled masked attention to keep individual identities distinct from cross-target interference, and sparse memory for stable temporal evolution, along with specialized strategies for occlusion handling and overlap prevention. SAM-MT successfully decouples latency from the number of targets, achieving real-time speed on par with single-target baselines ($ > $36 FPS for 10 targets) while maintaining SAM2's robust video segmentation performance.
}
\metadata[Code]{\url{https://github.com/FudanCVL/SAM-MT}}
\metadata[Email]{henghui.ding@gmail.com}

\begin{document}
\maketitle

\section{Introduction}

Contemporary Video Object Segmentation (VOS)~\cite{ding2023mose,MOSEv2} tracks and segments user-specified objects in open-vocabulary environments, with applications spanning in-the-wild navigation \cite{wang2025genie} and robotics \cite{griffin2020video}. Dominant approaches follow the space-time-memory (STM) paradigm \cite{oh2019video}, exploiting heavy pixel-level representations stored in a dense memory to track and segment target and have achieved state-of-the-art (SOTA) performance across VOS benchmarks \cite{oh2019video, cheng2021rethinking, cheng2022xmem, cheng2024putting, ravi2025sam, ding2025sam2long,ding2023mevis,ding2025mevis}.

\begin{figure*}[t]
  \centering
  \captionsetup[subfigure]{justification=centering, labelfont=normalfont}
  \begin{subfigure}[t]{0.28\textwidth}
    \centering
    \includegraphics[height=3cm, width=\textwidth]{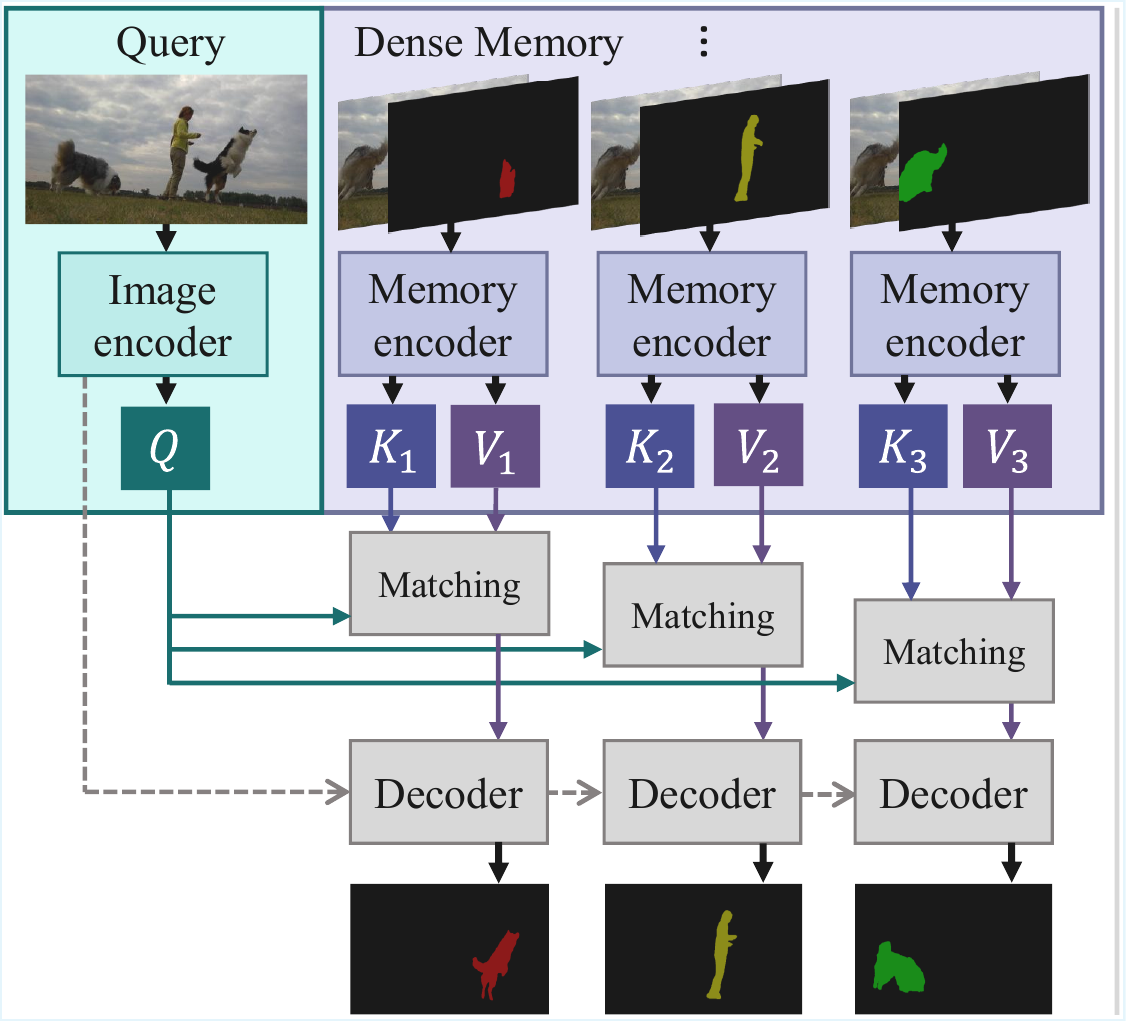}
    \captionsetup{justification=centering}
    \caption[STM and variants]{STM, SAM2, etc.\\ \cite{oh2019video, cheng2021modular, hu2021learning, xie2021efficient, wang2021swiftnet, ravi2025sam, ding2025sam2long}}
    \label{fig:arch_a}
  \end{subfigure}
  \hspace{-6pt}
  \begin{subfigure}[t]{0.32\textwidth}
    \centering
    \includegraphics[height=3cm, width=\textwidth]{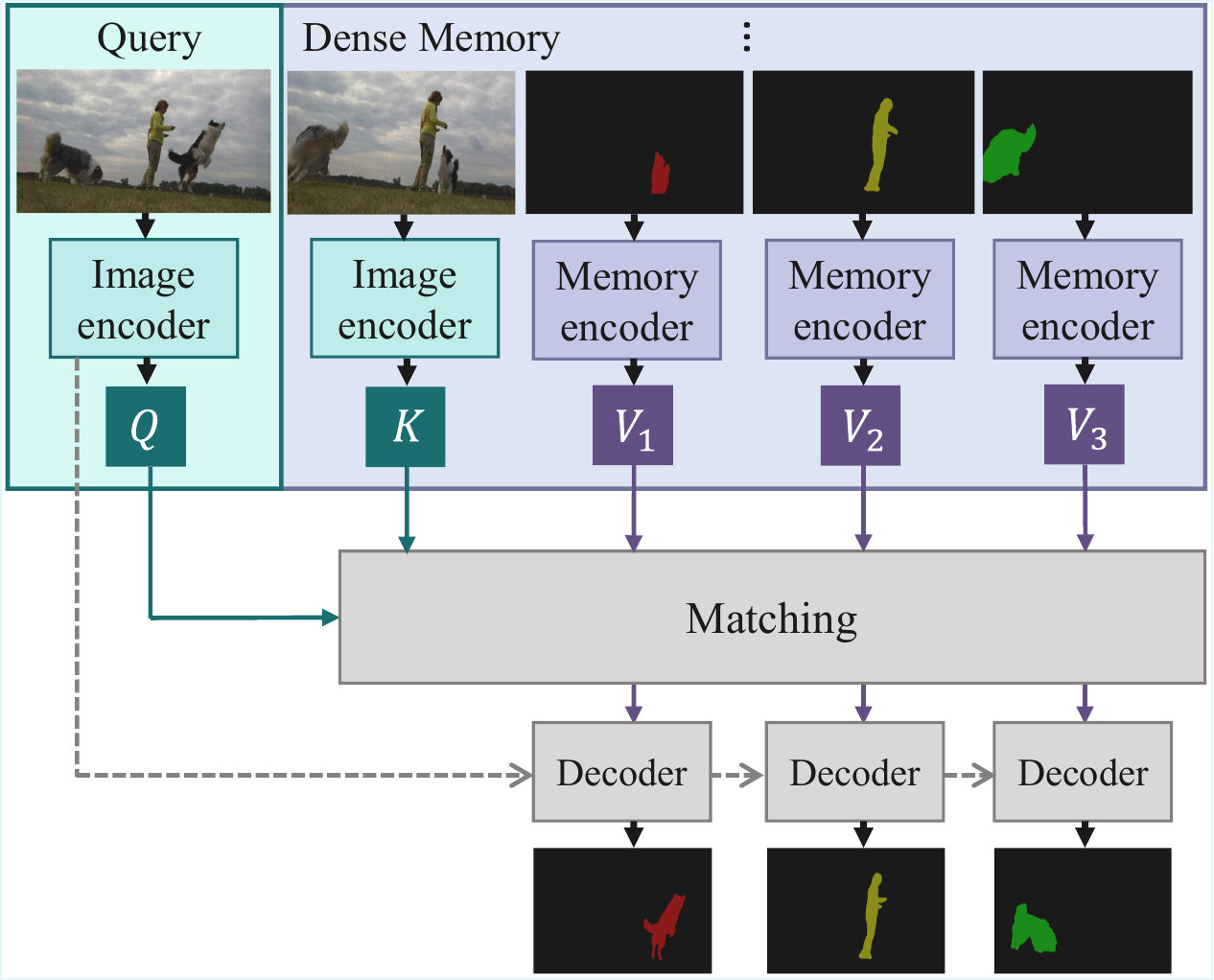}
    \captionsetup{justification=centering}
    \caption[STCN and variants]{STCN, XMem, Cutie, etc.\\ \cite{cheng2021rethinking, cheng2022xmem, bekuzarov2023xmem++, cheng2023tracking, cheng2024putting}}
    \label{fig:arch_b}
  \end{subfigure}
  \hspace{-6pt}
  \begin{subfigure}[t]{0.37\textwidth}
    \centering
    \includegraphics[height=2.99cm, width=\textwidth]{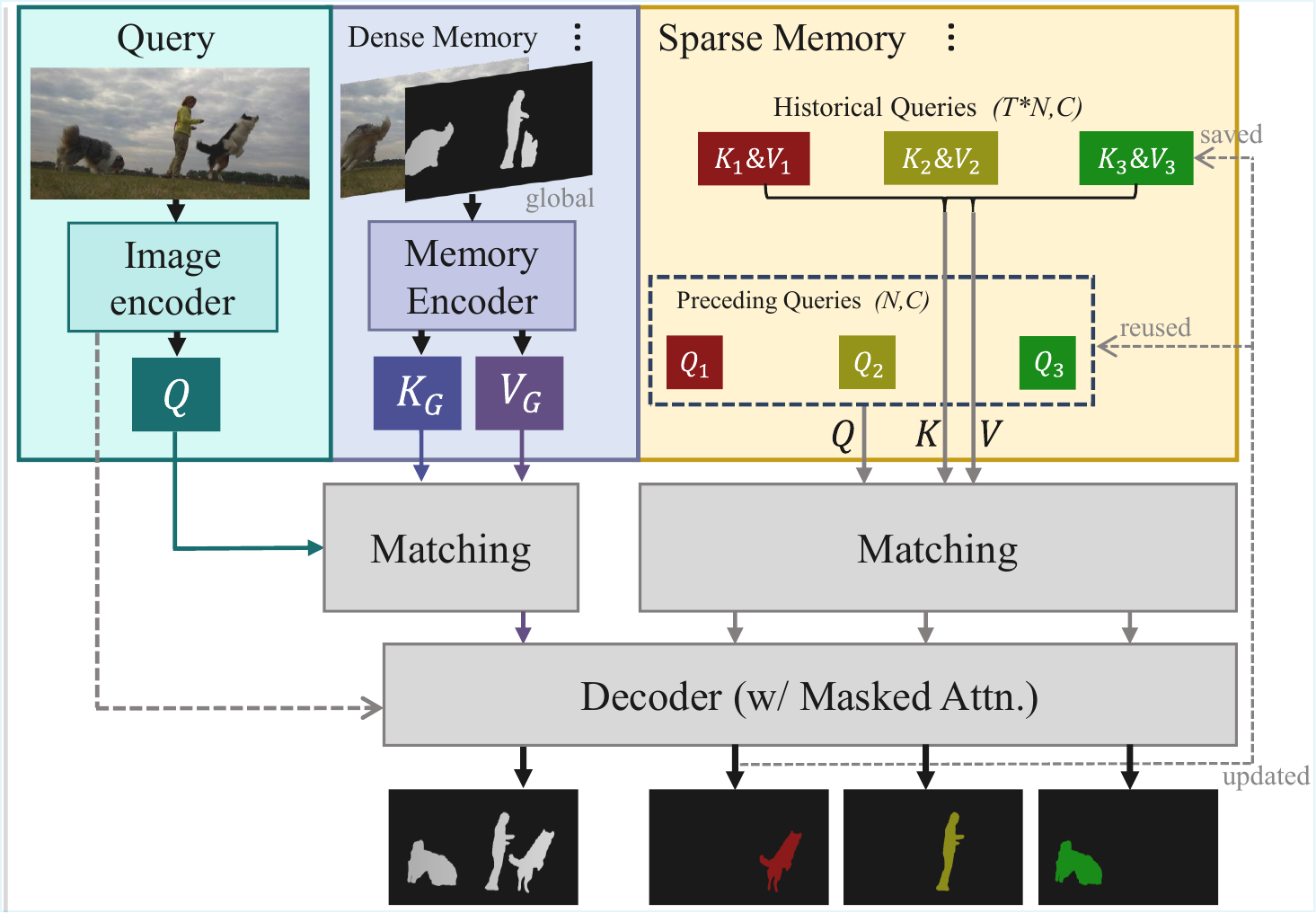}
    \captionsetup{justification=centering}
    \caption{SAM-MT (Ours)}
    \label{fig:arch_c}
  \end{subfigure}
  \caption{Comparison of video segmentation architectures: Single-target methods, implemented via (a) fully independent encoding or (b) feature sharing, still rely on object-wise processing, resulting in computational costs and latency to grow rapidly with the number of targets. In contrast, (c) SAM-MT models global context with a shared representation and individual targets with lightweight queries, achieving near-single-object efficiency as target count increases while maintaining SAM2's video segmentation performance.}
  \label{fig:architecture_compare}
  \vspace{-5pt}
\end{figure*}

While highly impressive, most of these approaches, including the powerful SAM2~\cite{ravi2025sam} and its extensions~\cite{ding2025sam2long, yang2024samurai, videnovic2025distractor}, rely on object-wise memory and propagation, making them tailored for single-target processing. Therefore, handling multiple targets requires repeating the same computations for every additional instance, as illustrated in Figure \ref{fig:architecture_compare}(a). Even with some earlier attempts to share frame-level image features, they still rely on independent object-wise processing for mask decoding and memory encoding \cite{cheng2021rethinking, cheng2022xmem, cheng2024putting} (Figure \ref{fig:architecture_compare}(b)). Consequently, such a straightforward extension results in a substantial rise in computational cost as the number of targets increases, leading to degraded frame rates (FPS) with unbounded latency. One possible workaround is to merge all targets into a single trackable object. However, this approach discards individual target IDs, and is therefore impractical for real-world applications. Furthermore, the merged object often forms irregular shapes that standard VOS models, trained on real-world objects, simply fail to recognize and track (see Section \ref{sec:qualitative evaluation}).

Meanwhile, real-world applications demand real-time tracking and segmentation of multiple targets. For instance, a practical VOS system for autonomous driving must 1) preserve individual target identities during propagation while 2) maintaining real-time speed, even in crowded scenes with dense vehicles and pedestrians. These requirements highlight the need for a real-time multi-target video segmentation framework that could \textit{maintain near-single-object efficiency as target count increases}, while faithfully \textit{preserving and updating individual identities} during propagation. Recent approaches, including SAM2, struggle to bridge this gap, as their object-wise processing causes computation to grow with target count.

We therefore present \textit{SAM-MT} to address this. As illustrated in Figure \ref{fig:architecture_compare}(c), SAM-MT extends the SAM2 architecture with a hybrid approach: it employs a shared representation to model the global context of all targets, while introducing \textit{target queries} in parallel for representing individual targets. To prevent cross-target interference, we propose \textit{decoupled masked attention}, which restricts attention between queries of different targets, while ensuring their shared access to the global context. For temporal evolution, a \textit{query-based sparse memory} stores historical queries from individual targets across frames for per-target re-identification. No extra mask decoders or memory encoders are required, allowing SAM-MT to avoid redundant object-wise computation and maintain near-single-object efficiency as the number of targets increases. For training, we adopt strided frame sampling rather than adjacent frames to better handle occlusions under GPU constraints. To further encourage instance-level separation, specialized supervision is introduced to penalize overlaps between different targets.

Comprehensive experiments demonstrate that SAM-MT achieves strong performance in real-world complex scenarios involving dense targets, frequent occlusions, and long-term sequences. It also performs on par with SAM2.1-B+ across six VOS benchmarks, including the challenging MOSEv2 \cite{MOSEv2} and the long-term LVOSv2 \cite{hong2025lvos}. Remarkably, SAM-MT maintains near-single-object efficiency as target count increases, running at 36+ FPS in crowded scenes with 10 targets. This presents an efficiency advantage over prior methods such as SAM2.1-B+, whose frame rate drops sharply from 37 FPS for a single target to 17.8 and 12.4 FPS for 3 and 5 targets, respectively.

In summary, our contributions are summarized as follows:
\vspace{-1mm}
\begin{itemize}
    \item We present \textit{SAM-MT}, an efficient framework for real-time interactive multi-target video segmentation. By decoupling computational costs from target count, SAM-MT achieves \textit{near-single-object efficiency} as target number increases, while maintaining SAM2's robust video segmentation performance.

    \item We introduce \textit{target queries} to represent individual targets and \textit{decoupled masked attention} to prevent cross-target interference while sharing global context, thereby preserving target identities throughout the sequence.

    \item We introduce a lightweight \textit{query-based sparse memory} to capture the temporal evolution of individual targets, with strided sampling and overlap prevention for enhanced robustness.

    \item SAM-MT achieves competitive video segmentation performance across six challenging VOS benchmarks in the interactive setting, while achieving the best efficiency across target densities. For 10 targets, it runs at a real-time speed of 36+ FPS, a $6\times$ speedup over SAM2.1-B+.

\end{itemize}

\section{Related Work}

\textbf{The single-target bottleneck of VOS.} \ 
The pioneering Space-Time Memory (STM) paradigm is inherently designed for single-target VOS, tracking a specific object by matching its dense pixel-level memory against query frames~\cite{oh2019video, wang2021swiftnet, seong2021hierarchical, park2022per}. Extending such methods to multiple targets typically requires replicating the single-object pipeline for each target, causing the computational burden to grow with target count. To tackle this, subsequent methods like STCN \cite{cheng2021rethinking}, XMem \cite{cheng2022xmem}, and Cutie \cite{cheng2024putting} reuse frame-level image features to share affinity computations, but they still rely on independent mask decoding and memory encoding for each object. Despite the success of SAM-family methods in video segmentation \cite{ravi2025sam, ding2025sam2long}, they still follow an object-wise processing paradigm, requiring the single-object pipeline to be replicated for each target and thus making multi-target streaming costly.

\noindent\textbf{Query-based Segmentation.} \
Query-based transformers have become the dominant architecture for image and video segmentation~\cite{li2023transformer,wu2023open}, with established approaches \cite{VLT-TPAMI,ReferringSurvey,cheng2021per, ghiasi2022scaling, cheng2022masked, he2023fastinst, he2026survey,li2023mask, jain2023oneformer, kirillov2023segment, ke2023segment, ye2025entitysam, liu2023gres, ding2021vision, wu2023continual, zhu2025rethinking,liu2024primitivenet,GREx} surpassing traditional CNN-based baselines \cite{long2015fully, chen2017deeplab, he2017mask, cheng2020boundary}. These methods typically use learnable queries that cross-attend to pixel-level features to produce rich target representations \cite{vaswani2017attention}, followed by task-specific heads for mask prediction. Recent video segmentation methods propagate queries across frames, but these queries often lack explicit target identities, either serving as generic object slots~\cite{cheng2021mask2former} or foreground-background representations~\cite{cheng2024putting}. In contrast, SAM-MT assigns each user-specified target an ID-aware query for identity-consistent propagation across frames.

\section{Method}

\label{sec:method}
\vspace{-0.05in}

\subsection{Preliminaries: SAM2}
\vspace{-0.05in}
SAM2 \cite{ravi2025sam} extends the Segment Anything \cite{kirillov2023segment} paradigm to video by conditioning current-frame features on a dense pixel-level memory. At frame $t$, the image features $F_t \in \mathbb{R}^{HW \times C}$ are refined by cross-attending to the memory readout $M_{\text{dense}, t} \in \mathbb{R}^{T \times HW \times C}$:
\vspace{-0.1in}
\begin{equation}
\tilde{F}_{t} = \text{softmax}(F_{t} M_{\text{dense}, t}^{\top}) M_{\text{dense}, t} + F_{t}.
\vspace{-0.1in}
\end{equation}

The mask decoder then generates masks from a set of queries $\mathbf{Q}=[G;P]$, where $G$ and $P$ denote the global queries and encoded user prompts, respectively. These queries are first updated via self-attention and then cross-attend to $\tilde{F}_{t}$ for contextual refinement and mask prediction. The target is then encoded into a dense pixel-level memory representation, typically with $4096$ tokens, along with a compact object pointer as an auxiliary object-level representation.

Since SAM2 maintains independent target representations and dense memories for each object, multi-target tracking thus requires replicated object-wise propagation, causing latency and memory usage to grow rapidly with target count.

\subsection{Overview of SAM-MT}
\begin{figure*}[t]
    \centering
    \includegraphics[width=1.0\textwidth]{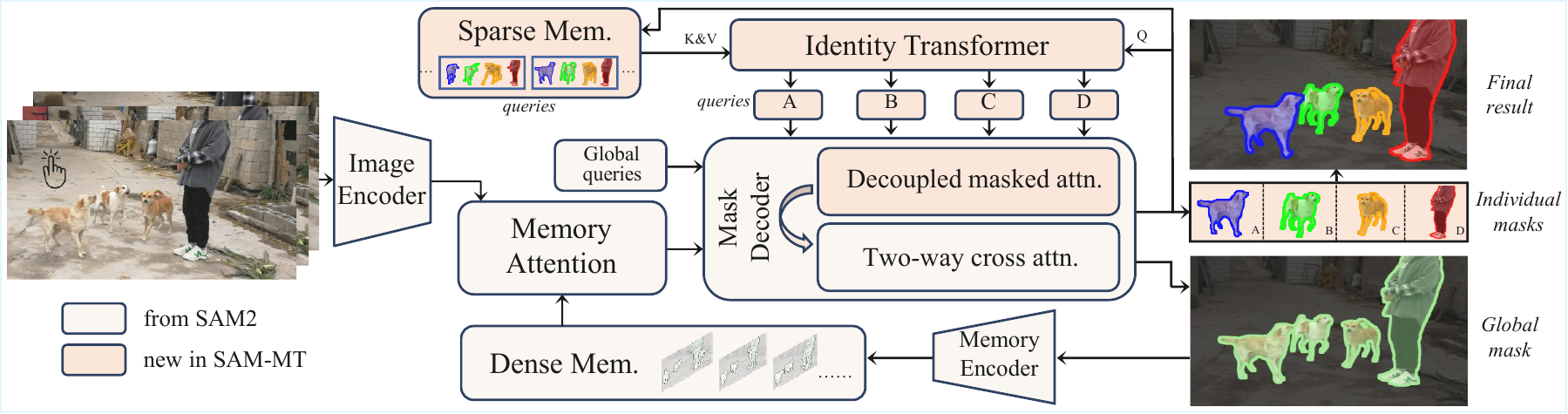}
    \caption{Overview of SAM-MT. Our framework uses queries to represent individual targets, in parallel with a shared representation for global context. We introduce \textit{decoupled masked attention} to prevent cross-target interference while maintaining access to the global context. A \textit{sparse memory} stores historical queries of different targets, which are processed by an \textit{identity transformer} for robust per-frame target re-identification.}
    \label{fig:method}
\end{figure*}

\vspace{-0.05in}
SAM-MT is a real-time multi-target video segmentation framework, as shown in Figure \ref{fig:method}. It adopts point-based interactions, where clicks are associated with target IDs to specify distinct objects.

The input points are encoded as queries and combined with SAM2's global queries in the decoder. To preserve individual target identities, we introduce \textit{decoupled masked attention}, which blocks cross-target interference while allowing each target to interact with the global context. The resulting queries then cross-attend to image features for pixel-level contextual refinement, following \cite{ravi2025sam}.

The refined point queries for each target are then consolidated into a unique target query (one query per target), which is used for mask prediction and stored in a unified first-in-first-out (FIFO) \textit{sparse memory} shared across targets. For subsequent frames, an \textit{identity transformer} retrieves the target-specific historical queries from the sparse memory to update the propagated target queries, which are then fed into the decoder to repeat the process, enabling consistent segmentation across frames.

In the following, we detail the core components of SAM-MT: decoupled masked attention, query-based sparse memory, and identity transformer. For intra-frame modules, temporal subscripts $t$ are omitted for brevity. We do not claim novelty for modules inherited from SAM2.

\subsection{Scalable Target Queries}
\vspace{-0.05in}
Existing query-based video segmentation methods typically allocate a fixed number of generic queries (e.g., 16 in~\cite{cheng2024putting}), making them difficult to scale to real-world scenes with arbitrary numbers of user-specified targets. As illustrated in Figure~\ref{fig:method}, SAM-MT adopts a scalable design that assigns explicit target queries to user-specified targets, naturally accommodating varying target counts.

\subsection{Decoupled Masked Attention}
\label{sec:decoupled_attn}
\vspace{-0.05in}
To prevent cross-target interference, we introduce a \textit{decoupled masked attention} strategy within the self-attention blocks of the decoder. Let $G \in \mathbb{R}^{N_g \times C}$ denote the global queries, $Q_q \in \mathbb{R}^{N_q \times C}$ denote the queries of target $q$, and $k$ denote the number of targets, we concatenate all queries as $\mathbf{Q} = [G; Q_1; \dots; Q_k] \in \mathbb{R}^{N \times C}$, where $N = N_g + \sum_{q=1}^k N_q$. Here, $N_q$ is the number of user-provided clicks for target $q$ in the initial frame, and simplifies to $N_q = 1$ for all subsequent frames as each target is represented by a single target query (see Section \ref{sec:Query Consolidation}). The decoupled masked attention with residual connection \cite{he2016deep} is then computed as:
\begin{equation}
\mathbf{Q}' = \text{softmax}(\mathbf{Q} K^\top + \mathcal{M}) V + \mathbf{Q}, 
\end{equation}
where \( \mathcal{M} \in \mathbb{R}^{N \times N }\) denotes the attention mask with entries:
\begin{equation}
\mathcal{M}_{i,j} =
\begin{cases}
0 & \text{if } i \in [1, N_g] \text{ and } j \in [1, N], \\
0 & \text{if } i \in [1, N] \text{ and } j \in [1, N_g], \\
0 & \text{if } i, j \in \text{Rng}(q), \quad \text{for } q \in \{1, \dots, k\}, \\
-\infty & \text{otherwise},
\end{cases}
\end{equation}
with \( \text{Rng}(q) \) defined as the index range of queries for target $q$:
\begin{equation}
\text{Rng}(q) = \left\{ N_g + \sum_{r=1}^{q-1} N_r + 1, \dots, N_g + \sum_{r=1}^{q} N_r \right\}.
\end{equation}

\begin{figure*}[!t]
    \centering
    \includegraphics[width=0.95\textwidth]{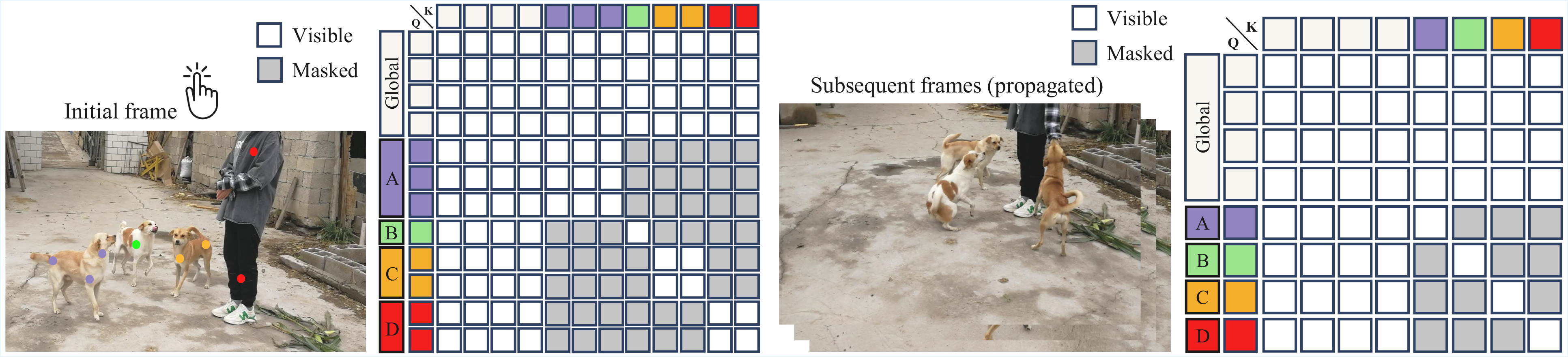}
    \caption{Visualization of decoupled masked attention. \textit{Left (Initial frame):} An arbitrary number of queries per target (e.g., 3, 1, 2, and 2 for targets A--D) are mutually isolated while sharing the global context. \textit{Right (Subsequent frames):} Propagated target queries (one query per target) follow the same masking rule.}
    \label{fig:decoupled masked attention}
\end{figure*}

The resulting \( \mathbf{Q}' \) aggregates the updated global and individual contexts, from which $Q'_q$ is extracted for target $q$ via $\text{Rng}(q)$. Figure \ref{fig:decoupled masked attention} illustrates the effect of decoupled masked attention in both initial and subsequent frames: interference between queries of different targets is blocked, while all target queries share access to the global context. Following this, all queries cross-attend to the image features $\tilde{F}_{t}$ as in SAM2~\cite{ravi2025sam}, extracting pixel-level context for mask prediction.

\subsection{Query Consolidation (Initial Frame)}
\label{sec:Query Consolidation}
\vspace{-1mm}
For each target \(q\), users may provide an arbitrary number of \( N_q \) points in the initial frame. These points are processed by the decoder into \( Q'_q \in \mathbb{R}^{N_q \times C} \) as described in Section \ref{sec:decoupled_attn}. To obtain a compact representation for the target, we apply a lightweight MLP-based weighting head \( f_{\text{weight}} \) to compute the importance \( w_q \in \mathbb{R}^{N_q} \) of each of its points, aggregating \( Q'_q \in \mathbb{R}^{N_q \times C} \) into a single query $\hat{Q}_q \in \mathbb{R}^{1 \times C}$, denoted as the \textit{target query}:
\vspace{-0.5em}
\begin{equation}
\vspace{-0.5em}
w_{q,i} = \text{softmax}(f_{\text{weight}}(Q'_{q,i})), \quad
\hat{Q_q} = \sum_{i=1}^{N_q} w_{q,i} \cdot Q'_{q,i}.
\end{equation}

This process is performed only in the initial frame, producing a unique target query for each target \(q\). For subsequent frames, since each target is represented by its target query that is propagated and updated frame by frame, consolidation is no longer needed and we simply have $\hat{Q_q} = Q'_{q}$.

\subsection{Query-based Sparse Memory}
As SAM2 relies on dense pixel-level memory $M_{\text{dense}, t} \in \mathbb{R}^{T \times HW \times C}$, duplicating such memory for each target is computationally prohibitive. To maintain real-time efficiency, we use dense memory only for the combined mask of all targets to capture global context. In parallel, we introduce a \textit{query-based sparse memory} to model the temporal evolution of individual targets.

Formally, let $\hat{Q}_{q,t} \in \mathbb{R}^{1 \times C}$ denote the updated query for target $q$ at frame $t$. We aggregate the queries of all $k$ targets into $\hat{\mathbf{Q}}_t = [\hat{Q}_{1,t}, \dots, \hat{Q}_{k,t}] \in \mathbb{R}^{k \times C}$ and update the FIFO sparse memory $M_{\text{sparse}, t}$, which maintains historical queries over a temporal window size of $T$ frames, including the initial frame and the $T-1$ most recent frames:
\begin{equation}
\vspace{-0.5em}
M_{\text{sparse}, t} = [\hat{\mathbf{Q}}_t, \hat{\mathbf{Q}}_{t-1}, \dots, \hat{\mathbf{Q}}_{t-T+2}, \hat{\mathbf{Q}}_{0}] \in \mathbb{R}^{T \cdot k \times C}.
\end{equation}

The sparse memory reduces the frame-level per-target storage cost from $HW$ dense tokens to one target-query token, thus enabling a much longer temporal window that could improve robustness to target occlusions and reappearances.

\subsection{Identity Transformer}
To maintain identity consistency across frames, we use an identity transformer to update target queries with their corresponding historical queries from the sparse memory. Specifically, at frame $t$, the target queries from the previous frame $\hat{\mathbf{Q}}_{t-1} \in \mathbb{R}^{k \times C}$ cross-attend to the sparse memory $M_{\text{sparse}, t-1} \in \mathbb{R}^{T \cdot k \times C}$. To prevent cross-target interference, an identity-aware mask $\tilde{\mathcal{M}}$ is applied within the update:
\begin{equation}
\mathbf{Q}_{t} = \text{softmax}(\hat{\mathbf{Q}}_{t-1} M_{\text{sparse}, t-1}^\top + \tilde{\mathcal{M}}) M_{\text{sparse}, t-1} + \hat{\mathbf{Q}}_{t-1},
\end{equation}
where the mask \( \tilde{\mathcal{M}} \in \mathbb{R}^{k \times (T \cdot k)} \) ensures that each target query only attends to its own history:
\begin{equation}
\tilde{\mathcal{M}}_{i,j} =
\begin{cases}
0 & \text{if } j \in \Omega_i, \\
-\infty & \text{otherwise},
\end{cases}
\end{equation}
with $\Omega_i = \{ i + \tau \cdot k \mid 0 \le \tau \le T-1 \}$ denoting the indices of historical queries for target $i$. 

By restricting each target to its own history, this design preserves identity consistency across frames and mitigates potential identity drift.

\section{Experimental Setup}

\subsection{Implementation Details}
\label{sec:implementation}
\noindent \textbf{Training Setup.} We initialize SAM-MT from the pre-trained SAM2.1-B+ \cite{ravi2025sam} checkpoint and train it on 8 NVIDIA A6000 GPUs. Following \cite{ravi2025sam}, we use a learning rate of $3 \times 10^{-6}$ for the image encoder and $5 \times 10^{-6}$ for other components. Training strategies are detailed in Section~\ref{sec:training_strategy}.

\noindent \textbf{Training Data.} We train SAM-MT on a filtered subset of the SA-V training set~\cite{ravi2025sam}, retaining sequences with at least three concurrent targets ($\approx$ 35\% of the original training set) to encourage robust multi-target handling.

\subsection{Training Strategies}
\label{sec:training_strategy}

\textbf{Training Scheme.} \ We employ a two-stage training scheme. Stage I focuses on static image-level training to ensure high-quality one-shot multi-target image segmentation from clicks. Stage II extends the training to video sequences, encouraging temporal identity consistency across frames.

\noindent \textbf{Strided Sampling.}
We sample 8 frames per sequence by combining consecutive sampling and 4-frame strided sampling (spanning a window of 32 frames). This enables the model to capture both short-term motion and long-term dynamics, such as occlusion and reappearance, under GPU memory constraints.

\noindent \textbf{Point Sampling.} To simulate realistic user interactions, we randomly sample $1$–$5$ positive and $0$-$2$ negative points per target in the initial frame of each sequence. We use a combination of grid-based sampling and random sampling to improve spatial coverage.

\noindent \textbf{Overlap Prevention.} 
To reduce spatial ambiguity and encourage each pixel only belongs to one object, we penalize mask overlaps between targets. Let $\mathbf{P}_i \in [0,1]^{H \times W}$ be the predicted probability map for target $i$, we define its overlap loss to encourage mutually exclusive predictions and alleviate pixel-level ambiguity:
\begin{equation} 
\mathcal{L}_{ovl, i} = \operatorname{Mean} \left( \mathbf{P}_i \odot \sum_{j \neq i} \mathbf{P}_j \right).
\end{equation}

\vspace{-0.05in}
\noindent \textbf{Loss Designs.}
The overall loss for SAM-MT consists of two parts: supervision for individual masks $\mathcal{L}_{I}$ and supervision for the global mask $\mathcal{L}_{G}$. For $k$ individual targets, we apply Focal and Dice losses together with the overlap loss to reduce identity ambiguity. For the global mask, we follow SAM2 \cite{ravi2025sam} and use Focal, Dice, IoU, and object score losses. Loss weights are omitted for brevity:
\vspace{-2mm}
\begin{equation}
\mathcal{L}_{I} = \frac{1}{k} \sum_{i=1}^{k} (\mathcal{L}_{focal, i} + \mathcal{L}_{dice, i} + \mathcal{L}_{ovl, i}),
\end{equation}
\vspace{-3mm}
\begin{equation}
\mathcal{L}_{G} = \mathcal{L}_{focal} + \mathcal{L}_{dice} + \mathcal{L}_{iou} + \mathcal{L}_{obj},
\end{equation}
\vspace{-3mm}
\begin{equation}
\mathcal{L}_{total} = \mathcal{L}_{I} + \mathcal{L}_{G}.
\end{equation}

This dual-level supervision encourages coherent global segmentation while improving the accuracy of individual target masks.

\subsection{Benchmarks and Metrics}
To evaluate SAM-MT's performance in complex real-world scenarios, we conduct quantitative evaluations on six challenging VOS benchmarks, including the validation splits of MOSEv2 \cite{MOSEv2}, MOSEv1 \cite{ding2023mose}, LVOSv2 \cite{hong2025lvos}, LVOSv1 \cite{hong2023lvos}, as well as SA-V val \cite{ravi2025sam} and SA-V test \cite{ravi2025sam}. Specifically, MOSE provides challenging cases such as frequent occlusions, rapid motion, low-light environments, and crowded scenes with inconspicuous targets, while LVOS focuses on long-term temporal consistency with object disappearances and reappearances. SA-V presents additional challenges in heavy occlusion and part-level object modeling. Following \cite{cheng2024putting, ravi2025sam, MOSEv2}, we report $\mathcal{J}\&\mathcal{F}$ for all benchmarks with additional $\mathcal{J}\&\dot{\mathcal{F}}$ for MOSEv2. For multi-target scalability analysis, we report Frames Per Second (FPS) to measure computational efficiency with respect to the number of targets, and VRAM to quantify GPU memory consumption.

\subsection{Initialization}
\label{sec:initialization}
\vspace{-1mm}
For a fair comparison, both SAM-MT and SAM2 are initialized with the same clicks, specifically two positive clicks per target in the initial frame of each sequence. In contrast, all other baselines are initialized with corresponding ground-truth masks, providing stronger initialization than our interactive setting.

\section{Experiments}
\label{sec:Eperiments}

\subsection{Quantitative Evaluation}
As shown in Table \ref{tab:main_quantitative_results}, SAM-MT achieves strong video segmentation performance across VOS benchmarks in a \textit{zero-shot} manner. Notably, on the challenging MOSEv2, SAM-MT reaches 43.0 $\mathcal{J}\&\dot{\mathcal{F}}$, improving over previous SOTA baselines including SAM2.1-B+ and Cutie. The gains are more pronounced in long-term scenarios, where SAM-MT surpasses SAM2.1-B+ by 2.0 and 2.3 points in $\mathcal{J}\&\mathcal{F}$ on LVOSv2 and LVOSv1, respectively. Table \ref{tab:sav_quantitative_results} further shows SAM-MT's robustness on SA-V, with competitive $\mathcal{J}\&\mathcal{F}$ on both splits.

\begin{table*}[!t]
\centering
\captionsetup{justification=centering}
\caption{\setstretch{0.4}Quantitative comparisons on VOS benchmarks. The best and second-best performances are marked in \colorbox{rred}{\vphantom{dg}red} and \colorbox{oorange}{\vphantom{dg}orange}, respectively. ``--'': unavailable due to out-of-memory.}
\label{tab:main_quantitative_results}
\vspace{-2mm}
\resizebox{\textwidth}{!}{
\renewcommand{\arraystretch}{1.1}
\begin{tabular}{l c ccccc c ccc c ccc c ccc}
\toprule
\multirow{2}{*}{Method} & \multirow{2}{*}{Initialization} & \multicolumn{5}{c}{MOSEv2-val} & & \multicolumn{3}{c}{MOSEv1-val} & & \multicolumn{3}{c}{LVOSv2-val} & & \multicolumn{3}{c}{LVOSv1-val} \\
\cmidrule(lr){3-7} \cmidrule(lr){9-11} \cmidrule(lr){13-15} \cmidrule(lr){17-19}
 & & $\mathcal{J}\&\dot{\mathcal{F}}$ & $\mathcal{J}$ & $\dot{\mathcal{F}}$ & $\mathcal{F}$ & $\mathcal{J}\&\mathcal{F}$ & & $\mathcal{J}\&\mathcal{F}$ & $\mathcal{J}$ & $\mathcal{F}$ & & $\mathcal{J}\&\mathcal{F}$ & $\mathcal{J}$ & $\mathcal{F}$ & & $\mathcal{J}\&\mathcal{F}$ & $\mathcal{J}$ & $\mathcal{F}$ \\ \midrule
STCN \cite{cheng2021rethinking} & mask & 29.7 & 28.9 & 30.5 & 31.4 & 30.2 & & 50.8 & 46.6 & 55.0 & & 65.3 & 61.6 & 68.9 & & 45.8 & 41.1 & 50.5 \\
AOT-L \cite{yang2021associating} & mask & 30.2 & 29.0 & 31.4 & 32.9 & 31.0 & & 57.2 & 53.1 & 61.3 & & 63.9 & 60.0 & 67.8 & & 59.4 & 53.6 & 65.2 \\
R50-AOT \cite{yang2021associating} & mask & 34.9 & 33.1 & 36.6 & 38.9 & 36.0 & & 58.5 & 54.4 & 62.6 & & \multicolumn{3}{c}{--} & & \multicolumn{3}{c}{--} \\
SwinB-AOT \cite{yang2021associating} & mask & 34.2 & 32.4 & 36.0 & 38.4 & 35.4 & & 60.2 & 56.2 & 64.1 & & \multicolumn{3}{c}{--} & & \multicolumn{3}{c}{--} \\
DeAOT-L \cite{yang2022decoupling} & mask & 32.6 & 30.7 & 34.5 & 37.2 & 33.9 & & 59.4 & 55.1 & 63.8 & & 67.0 & 62.4 & 71.6 & & 58.2 & 51.6 & 64.9 \\
R50-DeAOT \cite{yang2022decoupling} & mask & 31.6 & 29.7 & 33.5 & 36.1 & 32.9 & & 59.0 & 54.7 & 63.4 & & 71.4 & 67.8 & 75.1 & & 64.7 & 59.1 & 70.3 \\
SwinB-DeAOT \cite{yang2022decoupling} & mask & 36.7 & 34.9 & 38.6 & 41.0 & 37.9 & & 61.7 & 57.6 & 65.9 & & 72.6 & 69.0 & 76.2 & & 62.6 & 57.4 & 67.8 \\
RDE \cite{li2022recurrent} & mask & 32.0 & 30.7 & 33.3 & 35.0 & 32.8 & & 48.8 & 44.6 & 52.9 & & 61.5 & 58.0 & 65.0 & & 52.9 & 47.7 & 58.1 \\
XMem \cite{cheng2022xmem} & mask & 36.3 & 34.7 & 37.9 & 40.0 & 37.4 & & 57.6 & 53.3 & 62.0 & & 64.7 & 61.8 & 67.6 & & 50.0 & 45.5 & 54.4 \\
XMem++ \cite{bekuzarov2023xmem++} & mask & 34.2 & 32.5 & 35.9 & 37.9 & 35.2 & & 56.0 & 51.5 & 60.6 & & 61.5 & 58.6 & 64.3 & & 42.8 & 38.6 & 47.0 \\
DEVA \cite{cheng2023tracking} & mask & 38.3 & 36.6 & 40.0 & 42.2 & 39.4 & & 60.0 & 55.8 & 64.3 & & 72.4 & 68.7 & 76.1 & & 55.9 & 51.1 & 60.7 \\
Cutie-base \cite{cheng2024putting} & mask & \colorbox{oorange}{42.8} & \colorbox{oorange}{41.1} & \colorbox{oorange}{44.4} & \colorbox{oorange}{46.8} & \colorbox{oorange}{43.9} & & \colorbox{rred}{68.3} & \colorbox{oorange}{64.2} & \colorbox{rred}{72.3} & & 70.1 & 66.7 & 73.5 & & 66.0 & 61.3 & 70.6 \\ 
\midrule
SAM2.1-B+ \cite{ravi2025sam} & click & 41.1 & 39.4 & 42.8 & 45.4 & 42.4 & & 65.1 & 60.7 & 69.5 & & \colorbox{oorange}{74.6} & \colorbox{oorange}{70.9} & \colorbox{oorange}{78.3} & & \colorbox{oorange}{71.3} & \colorbox{oorange}{66.5} & \colorbox{oorange}{76.1} \\
\textbf{SAM-MT (Ours)} & click & \colorbox{rred}{43.0} & \colorbox{rred}{41.4} & \colorbox{rred}{44.5} & \colorbox{rred}{47.1} & \colorbox{rred}{44.2} & & \colorbox{oorange}{68.2} & \colorbox{rred}{64.4} & \colorbox{oorange}{72.0} & & \colorbox{rred}{76.6} & \colorbox{rred}{73.3} & \colorbox{rred}{80.0} & & \colorbox{rred}{73.6} & \colorbox{rred}{69.0} & \colorbox{rred}{78.3} \\ 
\bottomrule
\end{tabular}
}
\end{table*}

\begin{table*}[t]
\centering
\captionsetup{justification=centering}
\caption{Quantitative comparisons on SA-V validation and test splits.}
\vspace{-2mm}
\label{tab:sav_quantitative_results}
\resizebox{0.7\textwidth}{!}{
\scriptsize
\setlength{\fboxsep}{1.5pt}
\renewcommand{\arraystretch}{1.1}
\begin{tabular}{l c ccc c ccc}
\toprule
\multirow{2}{*}{Method} & \multirow{2}{*}{Initialization} & \multicolumn{3}{c}{SA-V val} & & \multicolumn{3}{c}{SA-V test} \\
\cmidrule(lr){3-5} \cmidrule(lr){7-9}
 & & $\mathcal{J}\&\mathcal{F}$ & $\mathcal{J}$ & $\mathcal{F}$ & & $\mathcal{J}\&\mathcal{F}$ & $\mathcal{J}$ & $\mathcal{F}$ \\ \midrule
SwinB-AOT \cite{yang2021associating} & mask & 51.1 & 46.4 & 55.7 & & 50.3 & 46.0 & 54.6 \\
SwinB-DeAOT \cite{yang2022decoupling} & mask & 61.4 & 56.6 & 66.2 & & 61.8 & 57.2 & \colorbox{oorange}{66.3} \\
RDE \cite{li2022recurrent} & mask & 51.8 & 48.4 & 55.2 & & 53.9 & 50.5 & 57.3 \\
XMem \cite{cheng2022xmem} & mask & 60.1 & 56.3 & 63.9 & & 62.3 & 58.9 & 65.8 \\
DEVA \cite{cheng2023tracking} & mask & 55.4 & 51.5 & 59.2 & & 56.2 & 52.4 & 60.1 \\
Cutie-base \cite{cheng2024putting} & mask & 60.7 & 57.7 & 63.7 & & 62.7 & 59.7 & 65.7 \\ \midrule
SAM2.1-B+ \cite{ravi2025sam} & click & \colorbox{oorange}{65.6} & \colorbox{oorange}{62.0} & \colorbox{oorange}{69.3} & & \colorbox{oorange}{66.1} & \colorbox{oorange}{62.4} & \colorbox{rred}{69.8} \\
\textbf{SAM-MT} & click & \colorbox{rred}{66.1} & \colorbox{rred}{62.9} & \colorbox{rred}{69.4} & & \colorbox{rred}{66.4} & \colorbox{rred}{63.0} & \colorbox{rred}{69.8} \\ \bottomrule
\end{tabular}
}
\vspace{-1mm}
\end{table*}

\subsection{Multi-target Scalability Analysis}
\begin{figure*}[!t]
  \centering
  \includegraphics[width=\linewidth]{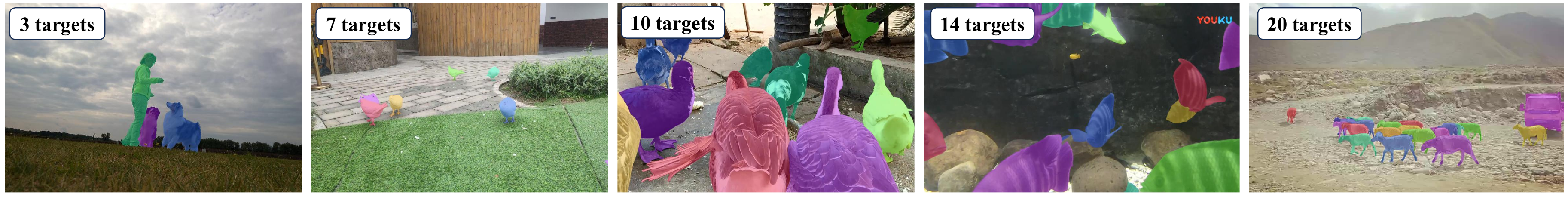}
  \vspace{-6mm}
  \captionsetup{justification=centering,singlelinecheck=false}
  \caption{Example videos from the multi-target synthetic benchmark (initial frames).}
  \label{fig:benchmark}
\end{figure*}

\textbf{Benchmark and Baselines.} Existing VOS datasets are dominated by single-target sequences (e.g., 79\% in MOSEv2-val). We therefore construct a synthetic benchmark by selecting multi-target sequences from those VOS datasets to assess multi-target scalability under varying numbers of targets. Specifically, the benchmark contains 20 sequences, corresponding to target counts from 1 to 20, with each sequence padded or truncated to 100 frames. Example sequences are shown in Figure~\ref{fig:benchmark}. We compare SAM-MT with SAM2.1-B+ \cite{ravi2025sam}, Cutie \cite{cheng2024putting}, as well as DeAOT \cite{yang2022decoupling}, a representative VOS method that also aims to decouple latency from target count. For a fair comparison, SAM2.1-B+ is evaluated with its official multi-target setting, including batch-wise target concatenation and feature reuse, where image features are computed once and shared across targets for acceleration. Cutie and DeAOT include these optimizations by design. All experiments are conducted on a single NVIDIA A6000 GPU with 48GB VRAM.

\begin{figure*}[!t]
\centering
\begin{minipage}[t]{0.60\linewidth}
  \centering
  \captionsetup{justification=centering,singlelinecheck=false}
  \captionof{table}{FPS comparison on the synthetic benchmark.}
  \label{tab:efficiency_comparison}
  \setlength{\tabcolsep}{3pt}
  \scriptsize
  \vspace{-3mm}
  \renewcommand{\arraystretch}{1.3}
  \resizebox{\linewidth}{!}{
  \begin{tabular}{lccccccccccc}
  \toprule
  \multirow{2}{*}{Method} & \multirow{2}{*}{Res.} & \multicolumn{9}{c}{FPS vs. Number of Targets} \\
  \cmidrule(lr){3-11}
   & & 1 & 2 & 3 & 5 & 7 & 9 & 11 & 15 & 20 \\
  \midrule
  DeAOT-L \cite{yang2022decoupling} & $1.3\times480p$ & 24.7 & 24.7 & 24.7 & 24.2 & 24.2 & 24.1 & 14.1 & 14.0 & 13.7 \\
  SwinB-DeAOT \cite{yang2022decoupling} & $1.3\times480p$ & 12.9 & 12.9 & 12.9 & 12.7 & 12.7 & 12.7 & \ 9.2 & \ 9.1 & \ 9.0 \\
  Cutie-base \cite{cheng2024putting} & $480p$ & 31.2 & 31.1 & 30.4 & 27.4 & 23.4 & 21.8 & 20.6 & 18.4 & 15.0 \\
  Cutie-base \cite{cheng2024putting} & $1024p$ & 20.3 & 17.5 & 14.7 & 13.2 & 11.2 & \ 9.4 & \ 8.9 & \ 7.3 & \ 6.3 \\
  \midrule
  SAM2.1-B+ \cite{ravi2025sam} & $1024p$ & \textbf{37.2} & 22.9 & 17.8 & 12.4 & \ 9.5 & \ 7.8 & \ 6.5 & \ 4.9 & \ 3.7 \\
  \rowcolor{defaultColor} \textbf{SAM-MT} & $1024p$ & \textbf{37.2} & \textbf{36.8} & \textbf{36.8} & \textbf{36.5} & \textbf{36.4} & \textbf{36.3} & \textbf{36.0} & \textbf{35.7} & \textbf{35.4} \\
  \bottomrule
  \end{tabular}
  }
\end{minipage}
\hfill
\begin{minipage}[t]{0.38\linewidth}
  \centering
  \raisebox{-21mm}{%
    \begin{minipage}{\linewidth}
      \centering
      \includegraphics[width=\linewidth,height=36mm,keepaspectratio]{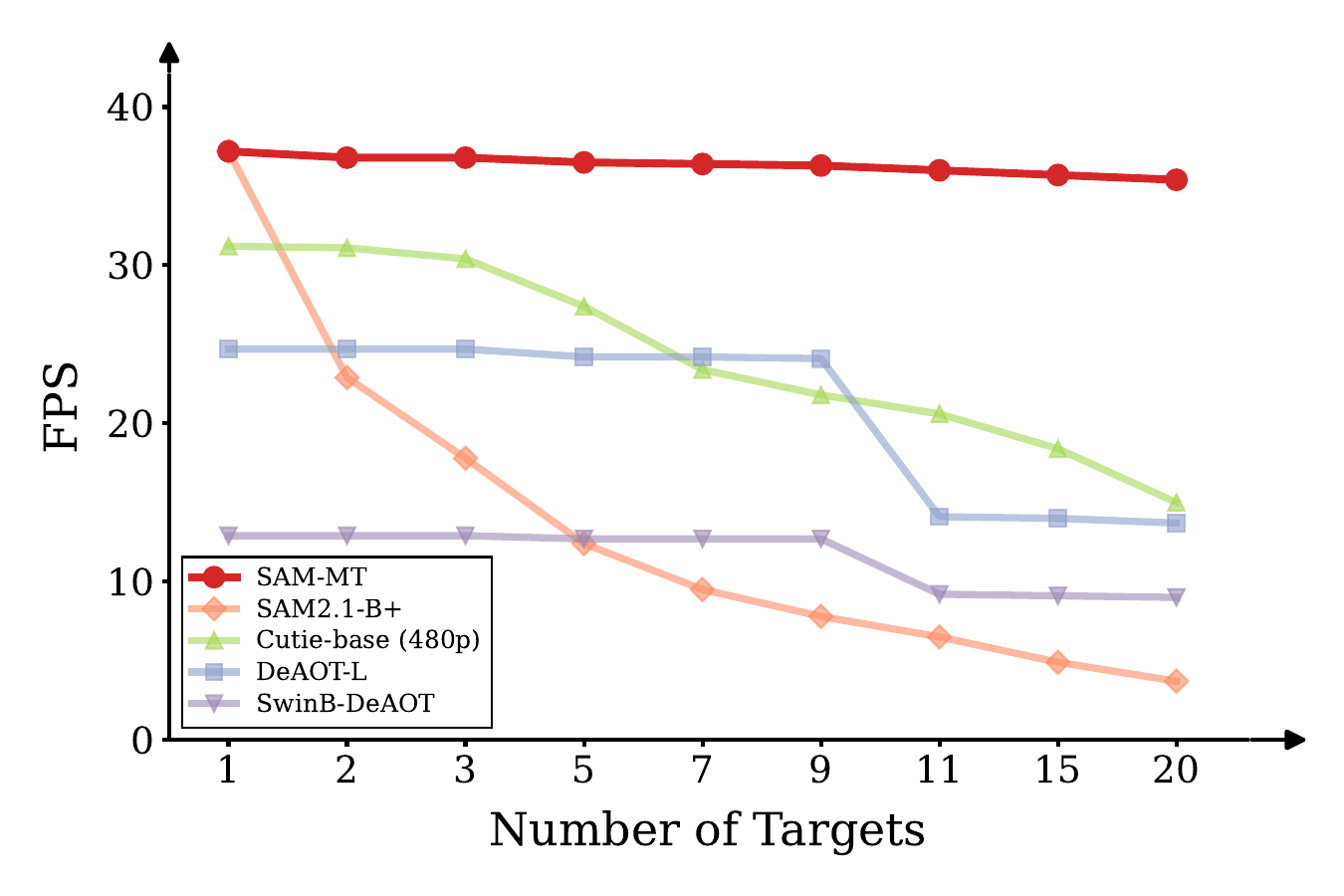}
      \vspace{-3mm}
      \captionsetup{font=scriptsize,justification=centering,singlelinecheck=false}
      \captionof{figure}{FPS on the synthetic benchmark.}
      \label{fig:multi_target_fps}
    \end{minipage}
  }
\end{minipage}
\end{figure*}

\noindent \textbf{FPS Comparison.}
As shown in Table~\ref{tab:efficiency_comparison} and Figure~\ref{fig:multi_target_fps}, SAM-MT largely decouples latency from target count, maintaining near-single-object efficiency as target density scales. It achieves the highest FPS across all target counts, despite processing higher-resolution inputs (1024p) than many baselines (480p). In contrast, both SAM2.1-B+ and Cutie suffer significant FPS degradation as the number of targets increases; for example, SAM2.1-B+ drops from 37.2 FPS (1 target) to 17.8 FPS (3 targets). Although DeAOT-L and SwinB-DeAOT maintain stable processing speeds for up to 10 targets, their speeds drop sharply beyond this point, and their overall frame rates remain much lower than ours.

\noindent \textbf{VRAM Comparison.}
Table~\ref{tab:vram_usage} compares VRAM usage under different target counts. SAM-MT keeps memory consumption nearly stable as target count increases, rising only from 3094 MB (1 target) to 3785 MB (20 targets). In contrast, SAM2.1-B+ grows substantially from 3043 MB to 8585 MB, showing the memory efficiency of our design in dense multi-target scenes.

\begin{table}[!t]
\centering
\scriptsize
\captionsetup{justification=centering,singlelinecheck=false}
\caption{VRAM (MB) comparison on the synthetic benchmark.}
\label{tab:vram_usage}
\vspace{-2mm}
\renewcommand{\arraystretch}{1.3}
\begin{tabular}{l|cccccccccc}
\toprule
Method / Target Num. & 1 & 3 & 5 & 7 & 9 & 11 & 13 & 15 & 17 & 20 \\
\midrule
SAM2.1-B+ & \textbf{3043} & 3356 & 3702 & 4089 & 4387 & 4529 & 4873 & 5436 & 6831 & 8585 \\
\rowcolor{blue!10}
\textbf{SAM-MT} & 3094 & \textbf{3247} & \textbf{3312} & \textbf{3357} & \textbf{3430} & \textbf{3491} & \textbf{3548} & \textbf{3627} & \textbf{3719} & \textbf{3785} \\
\bottomrule
\end{tabular}
\vspace{-2mm}
\end{table}

\noindent \textbf{FPS Comparison on VOS Benchmarks.}
Table~\ref{tab:vos_efficiency} reports FPS on VOS benchmarks under different target densities. Although these VOS benchmarks are dominated by single-target sequences, SAM-MT consistently maintains real-time speed with only minor degradation as target count increases (e.g., 36.9$\rightarrow$35.8 FPS on MOSEv2 for $\ge5$ targets). In contrast, SAM2.1-B+ degrades sharply from 32.1 to 11.5 FPS under the same setting, and Cutie shows a similar trend. While DeAOT remains relatively stable, its overall FPS is much lower.

\begin{table*}[!h]
  \centering
  \captionsetup{justification=centering,singlelinecheck=false}
  \caption{FPS comparison on VOS benchmarks. ``All'': entire validation set; ``$\ge$2'': subset containing sequences with $\ge$ 2 concurrent targets, etc.}
  \label{tab:vos_efficiency}
  \vspace{-3mm}
  \resizebox{\textwidth}{!}{
  \renewcommand{\arraystretch}{1.1}
  \begin{tabular}{lccccccccccccc}
  \toprule
  Method & Res. & \multicolumn{4}{c}{MOSEv2} & \multicolumn{4}{c}{MOSEv1} & \multicolumn{2}{c}{LVOSv2} & \multicolumn{2}{c}{LVOSv1} \\
  \cmidrule(l{0.2em}r{0.2em}){3-6} \cmidrule(l{0.2em}r{0.2em}){7-10} \cmidrule(l{0.2em}r{0.2em}){11-12} \cmidrule(l{0.2em}r{0.2em}){13-14}
   & & All & $\ge$2 & $\ge$3 & $\ge$5 & All & $\ge$2 & $\ge$3 & $\ge$5 & All & $\ge$2 & All & $\ge$2 \\
  \midrule
  DeAOT-L \cite{yang2022decoupling} & $1.3\times480p$ & 22.9 & 21.5 & 21.4 & 21.1 & 29.7 & 29.0 & 28.9 & 28.5 & \ 9.5 & \ 9.5 & \ 7.2 & \ 6.5 \\
  SwinB-DeAOT \cite{yang2022decoupling} & $1.3\times480p$& 12.7 & 12.7 & 12.6 & 12.3 & 15.8 & 15.8 & 15.6 & 14.6 & \ 6.9 & \ 6.7 & \ 5.6 & \ 4.9 \\
  Cutie-base \cite{cheng2024putting} & $480p$ & 44.3 & 41.0 & 37.9 & 34.8 & 42.7 & 35.9 & 34.6 & 27.7 & 43.4 & 41.9 & 44.8 & 39.6 \\
  Cutie-base \cite{cheng2024putting} & $1024p$ & 21.3 & 16.4 & 14.1 & 10.2 & 21.9 & 15.6 & 12.6 & 10.0 & 19.3 & 15.2 & 21.1 & 16.2 \\
  \midrule
  SAM2.1-B+ \cite{ravi2025sam} & $1024p$ & 32.1 & 21.3 & 17.3 & 11.5 & 30.5 & 19.0 & 14.5 & 10.8 & 32.0 & 25.6 & 32.2 & 22.1 \\
  \rowcolor{defaultColor} \textbf{SAM-MT} & $1024p$ & 36.9 & 36.2 & 36.1 & 35.8 & 39.2 & 37.1 & 36.6 & 36.1 & 36.5 & 35.6 & 36.7 & 36.0 \\
  \bottomrule
  \end{tabular}
  }
  \vspace{-3mm}
\end{table*}

\subsection{Qualitative Evaluation}
\label{sec:qualitative evaluation}
\begin{figure*}[!t]
    \centering
    \includegraphics[width=1.0\textwidth]{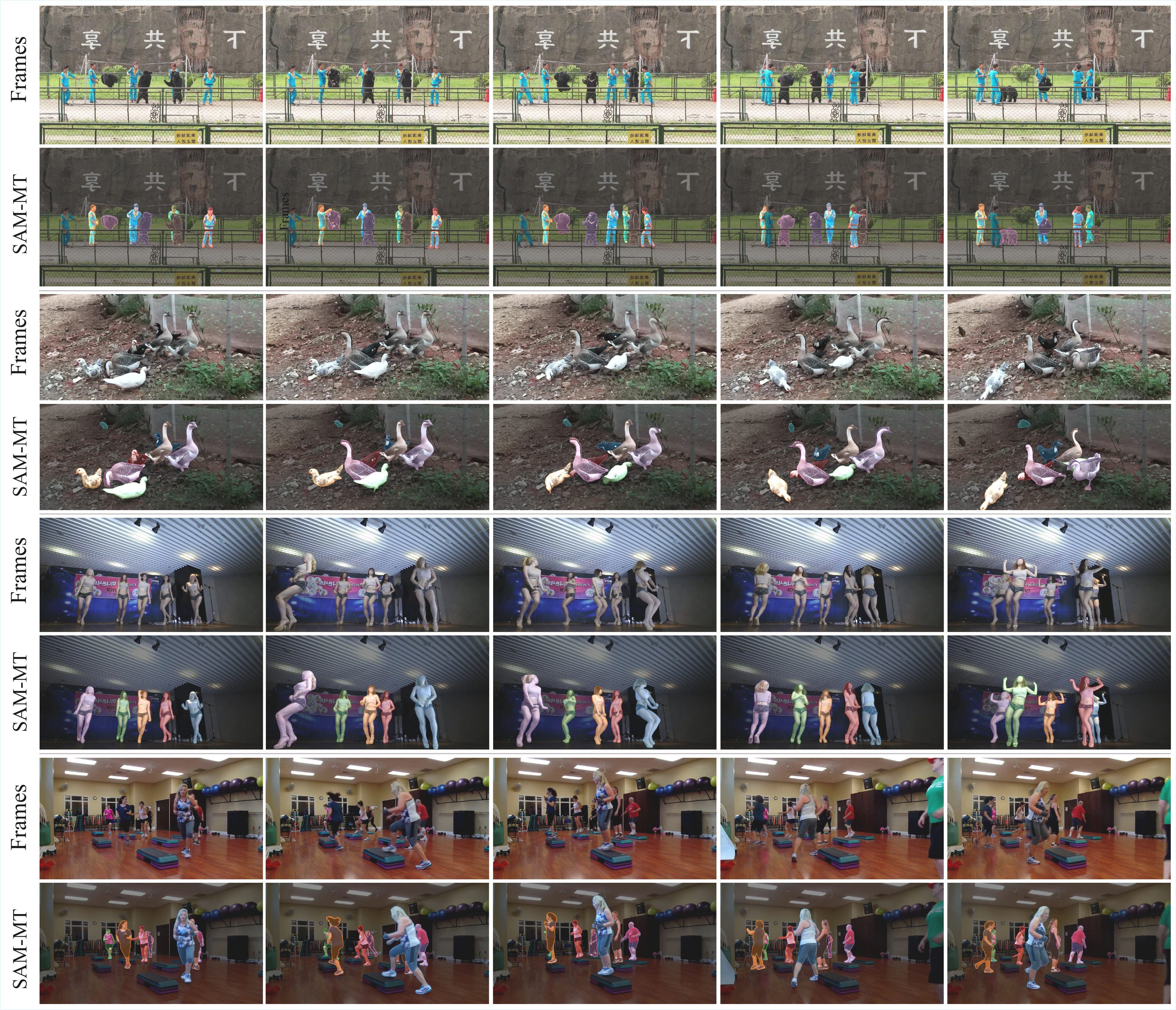}
    \vspace{-6mm}
    \caption{Qualitative results of SAM-MT in highly dense real-world scenarios. Our method achieves precise tracking and segmentation for all targets while maintaining real-time, near-single-object processing speed.}
    \label{fig:vis1}
\end{figure*}

\vspace{-0.05in}
\noindent \textbf{Highly Dense Scenarios.}
We evaluate SAM-MT in dense multi-target scenarios. As illustrated in Figure~\ref{fig:vis1}, SAM-MT achieves precise and consistent video segmentation in complex scenes, including a circus performance, a flock of ducks, and a crowded group fitness scene, demonstrating its robustness in maintaining and updating distinct identities in highly cluttered environments.

\begin{figure*}[!t]
    \centering
    \includegraphics[width=1.0\textwidth]{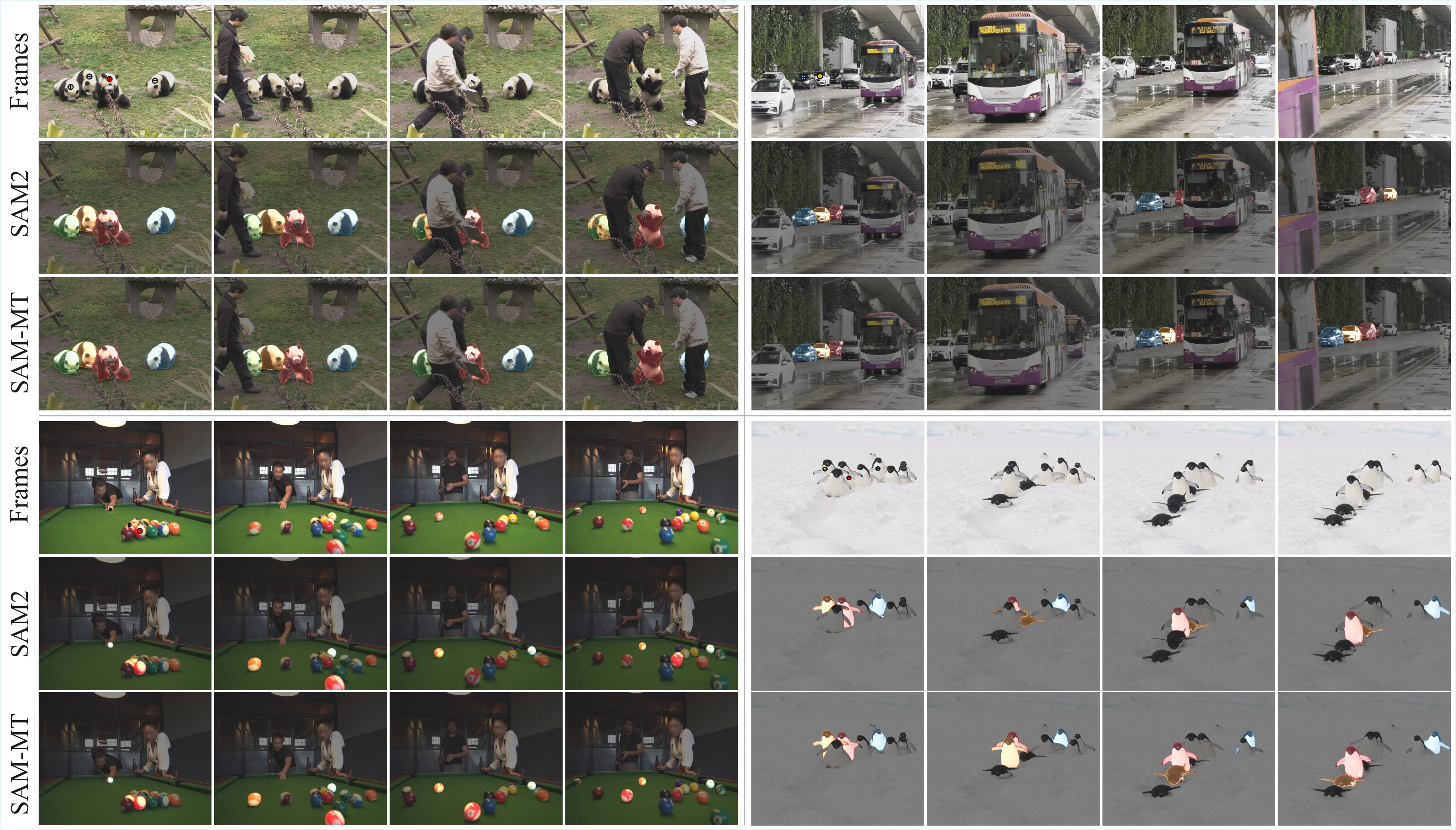}
    \vspace{-6mm}
    \caption{Qualitative comparisons in identity-ambiguous scenarios with multiple visually similar targets. SAM-MT maintains robust temporal identity consistency, whereas SAM2 often suffers from tracking loss or identity swaps.}
    \label{fig:vis2}
\end{figure*}

\noindent \textbf{Identity-Ambiguous Scenarios.}
SAM-MT combines global context and individual identities through decoupled masked attention, while SAM2 processes each target in isolation and may be affected by surrounding distractors. As shown in Figure~\ref{fig:vis2}, in crowded scenes with visually similar objects (e.g., pandas, vehicles, billiards, and penguins), SAM2.1-B+ tracks well initially but later suffers from either tracking loss or identity drift. In contrast, SAM-MT resolves these ambiguities and maintains consistent identities for all targets throughout.

\begin{figure*}[!t]
    \centering
    \includegraphics[width=1.0\textwidth]{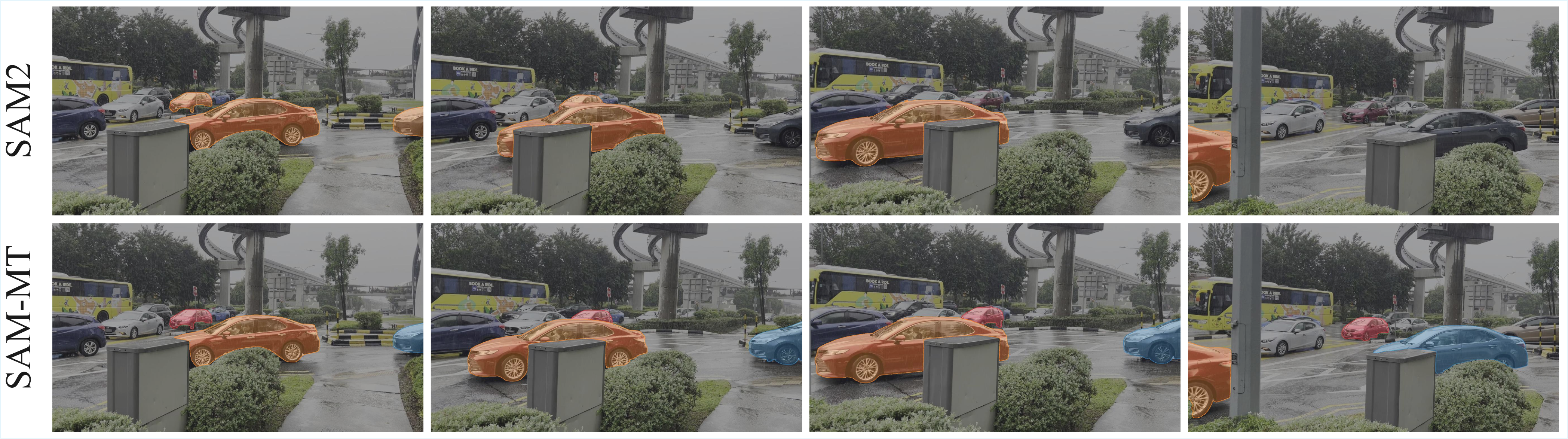}
    \vspace{-6mm}
    \caption{To maintain a single-object efficiency budget, attempting to track the merged target causes SAM2 to easily lose track. In contrast, SAM-MT precisely tracks and segments individual targets with well-preserved IDs under the same efficiency budget.}
    \label{fig:vis3}
\end{figure*}

\noindent \textbf{Merging Targets into a Single Mask in SAM2.}
For VOS models optimized for single-target propagation, such as SAM2, a simple workaround for maintaining real-time single-object efficiency under multiple targets is to merge all targets into a unified mask. However, this not only discards individual IDs but also forms an unnatural shape, making the ``target'' difficult to track. As shown in Figure~\ref{fig:vis3}, merging three cars into one fails to form a meaningful object and thus leads to complete tracking loss. In contrast, SAM-MT tracks and segments individual targets accurately at real-time, near-single-object speed.

\subsection{Ablation Studies}
\vspace{-1mm}
We conduct ablation studies on the MOSEv2 validation set~\cite{MOSEv2}. Following~\cite{MOSEv2}, we report \( \mathcal{J}\&\dot{\mathcal{F}} \) for video segmentation accuracy and FPS for multi-target scalability when applicable. For a fair comparison, all ablation variants are trained with the same protocols described in Sections~\ref{sec:implementation}, \ref{sec:training_strategy}, and~\ref{sec:initialization}.

\noindent \textbf{Ablation on Masked Attention Strategies.} \
We compare \textit{decoupled masked attention} with two alternatives: (a) Full Visibility, where all target queries freely interact with each other, and (b) Full Masking, where global queries are completely isolated from target queries. As shown in Table~\ref{tab:ablation_masked_attention}(a), allowing unrestricted query interactions causes target queries to lose their identity specificity, leading to a 5.5-point drop in $\mathcal{J}\&\dot{\mathcal{F}}$ from 43.0 to 37.5. In contrast, Table~\ref{tab:ablation_masked_attention}(b) shows that isolating global context from target queries weakens global-individual information exchange, resulting in a 3.7-point decrease in $\mathcal{J}\&\dot{\mathcal{F}}$. These results validate the rationale behind our decoupled design, which preserves target-specific identities while enabling effective interactions with global context.

\noindent \textbf{Ablation on Identity Transformer.} \
Table \ref{tab:ablation_identity_transformer} evaluates the identity-aware mask and transformer depth. Without the mask, each query attends to historical queries from all targets, causing cross-target memory pollution and a 3.9-point drop in $\mathcal{J}\&\dot{\mathcal{F}}$. Meanwhile, adding more transformer blocks improves accuracy but lowers FPS. We choose 3 blocks to balance speed and accuracy.

\noindent \textbf{Ablation on Window Size of Sparse Memory.} \
Table \ref{tab:ablation_window_size} evaluates the sparse-memory window size, where our lightweight queries enable SAM-MT to use a longer memory window than SAM2. As shown, increasing the window size from 8 to 32 improves $\mathcal{J}\&\dot{\mathcal{F}}$ by 0.7 points, but reduces FPS by 0.5 due to extra computation. We set window size to 16 for a better accuracy-speed trade-off.

\begin{figure*}[t]
    \centering
    \captionsetup{justification=centering,singlelinecheck=false}
    \includegraphics[width=1.0\textwidth]{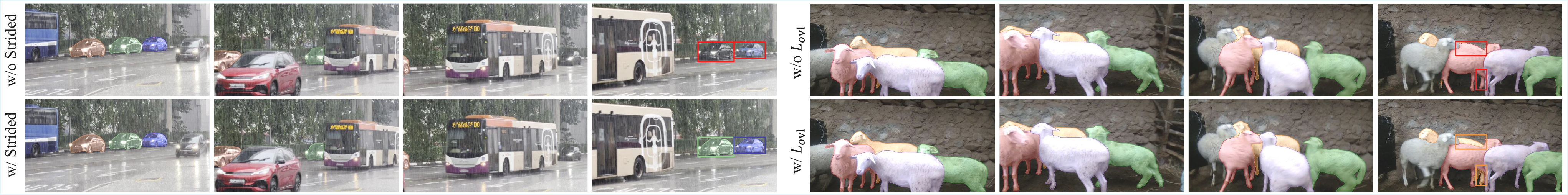}
    \vspace{-6mm}
    \caption{\textit{Left:} Strided sampling improves robustness to target reappearance. \textit{Right:} Overlap prevention reduces pixel-level identity ambiguity in dense scenes.}
    \label{fig:strided_overlap}
\end{figure*}

\noindent \textbf{Ablation on Training Strategies.} \
Table \ref{tab:ablation_training_strategies} evaluates the training strategies. (a) Removing static image pretraining results in a 2.7-point drop in $\mathcal{J}\&\dot{\mathcal{F}}$, showing the necessity of image-level pretraining for one-shot multi-target image segmentation. (b) Without strided sampling (i.e., only using adjacent frames as in SAM2 \cite{ravi2025sam}), $\mathcal{J}\&\dot{\mathcal{F}}$ decreases by 1.3 points; as shown in Figure \ref{fig:strided_overlap} (left), this often causes identity loss upon target reappearance. (c) As shown in Figure \ref{fig:strided_overlap} (right), overlap prevention alleviates mask conflicts in dense scenes (e.g., a group of moving sheep), ensuring clearer separation of individual targets.

\begin{table}[!t]
  \centering
  \scriptsize
  \captionsetup{font=scriptsize}
  \begin{minipage}[t]{0.49\textwidth}
    \centering
    \captionsetup{justification=centering,singlelinecheck=false}
    \caption{Ablation on Masked Attn. Strategies.}
    \label{tab:ablation_masked_attention}
    \vspace{-1em}
    \setlength{\tabcolsep}{1mm} 
    \renewcommand{\arraystretch}{1.55} 
    \begin{tabular}{llcccc}
      \toprule
      & Strategy & $\mathcal{J}\&\dot{\mathcal{F}}$ & $\mathcal{J}$ & $\dot{\mathcal{F}}$ & FPS \\
      \midrule
      (a) & Full Visibility & 37.5 & 35.8 & 39.1 & \textbf{37.1} \\
      (b) & Full Masking  & 39.3 & 37.7 & 40.8 & 36.9 \\
      \rowcolor{defaultColor} (c) & \textbf{Decoupled} & \textbf{43.0} & \textbf{41.4} & \textbf{44.5} & 36.9 \\
      \bottomrule
    \end{tabular}
  \end{minipage}%
  \begin{minipage}[t]{0.51\textwidth}
    \centering
    \captionsetup{justification=centering,singlelinecheck=false}
    \caption{Ablation on Identity Transformer.}
    \label{tab:ablation_identity_transformer}
    \vspace{-1em}
    \setlength{\tabcolsep}{1mm} 
    \renewcommand{\arraystretch}{1.1}
    \begin{tabular}{lccccc}
      \toprule
      Strategy & Depth & $\mathcal{J}\&\dot{\mathcal{F}}$ & $\mathcal{J}$ & $\dot{\mathcal{F}}$ & FPS \\
      \midrule
      w/o Masking & 3 & 39.1 & 37.4 & 40.7 & 37.1 \\
      \midrule
      & 1 & 41.9 & 40.3 & 43.4 & \textbf{37.2} \\
      \rowcolor{defaultColor} \textbf{w/ Masking} & 3 & 43.0 & 41.4 & 44.5 & 36.9 \\
      & 5 & \textbf{43.3} & \textbf{41.6} & \textbf{45.0} & 36.4 \\
      \bottomrule
    \end{tabular}
    \vspace{0.7em}
  \end{minipage}
  \begin{minipage}[t]{0.49\textwidth}
    \centering
    \captionsetup{justification=centering,singlelinecheck=false}
    \caption{Ablation on Window Size.}
    \label{tab:ablation_window_size}
    \vspace{-1em}
    \setlength{\tabcolsep}{1.6mm} 
    \renewcommand{\arraystretch}{1.15} 
    \begin{tabular}{lccccc}
      \toprule
       & \makecell[c]{Window\\Size} & $\mathcal{J}\&\dot{\mathcal{F}}$ & $\mathcal{J}$ & $\dot{\mathcal{F}}$ & FPS \\
      \midrule
      (a) & 8  & 42.4 & 40.8 & 44.0 & \textbf{37.1} \\
      (b) & 12 & 42.6 & 41.0 & 44.2 & \textbf{37.1} \\
      \rowcolor{defaultColor} (c) & 16 & 43.0 & 41.4 & 44.5 & 36.9 \\
      (d) & \textbf{32} & \textbf{43.1} & \textbf{41.6} & \textbf{44.6} & 36.6 \\
      \bottomrule
    \end{tabular}
  \end{minipage}%
  \begin{minipage}[t]{0.51\textwidth}
    \centering
    \captionsetup{justification=centering,singlelinecheck=false}
    \caption{Ablation on Training Strategies.}
    \label{tab:ablation_training_strategies}
    \vspace{-1em}
    \setlength{\tabcolsep}{1.5mm} 
    \renewcommand{\arraystretch}{1.15}
    \begin{tabular}{ccccc}
      \toprule
       & \makecell[c]{Image\\Pretrain} & \makecell[c]{Strided\\Sampling} & \makecell[c]{Overlap\\Loss} & $\mathcal{J}\&\dot{\mathcal{F}}$ \\
      \midrule
      (a) &             & \checkmark & \checkmark & 40.3 \\
      (b) & \checkmark &            & \checkmark & 41.7 \\
      (c) & \checkmark & \checkmark &            & 42.5 \\
      \rowcolor{defaultColor} (d) & \checkmark & \checkmark & \checkmark & \textbf{43.0} \\
      \bottomrule
    \end{tabular}
  \end{minipage}
  \vspace{-3mm}
\end{table}

\subsection{Limitations.} 
To the best of our knowledge, SAM-MT is the first work to tackle the SAM family's multi-target bottleneck at the framework level, achieving near-single-object efficiency with competitive video segmentation accuracy. However, as it inherits SAM's vision-only architecture, it lacks the reasoning capability required for complex high-level tasks. Extending it with multimodal or reasoning \cite{lai2024lisa,weihua2026adamcot,weihua2025ccl,bai2024one, yuan2025sa2va,zheng2025mma} is a promising future direction.

\section{Conclusion}

We present SAM-MT, the first framework to tackle the multi-target bottleneck of the SAM family for real-time interactive video segmentation. SAM-MT represents individual targets using lightweight queries in parallel with shared global context, enabling near-single-object efficiency as the number of targets increases. Extensive experiments show that SAM-MT maintains SAM2's strong segmentation performance across VOS benchmarks, while achieving real-time speed and robust performance in dense in-the-wild scenarios. We hope SAM-MT represents a solid step toward real-time interactive multi-target video segmentation.

\bibliographystyle{unsrtnat}
\bibliography{mybib.bib}

\end{document}